\documentclass[11pt,a4paper]{article}
\usepackage[hyperref]{emnlp2020}
\usepackage{times}
\usepackage{latexsym}

\usepackage{graphicx}

\usepackage{makecell}
\usepackage{amsmath}
\usepackage{booktabs}
\usepackage{float}

\usepackage{microtype}

\aclfinalcopy 

\newcommand{\emldisplay}[2]{\texttt{\href{mailto:#1}{#2}}}
\newcommand{\eml}[1]{\emldisplay{#1}{#1}}

\title{EdinburghNLP at WNUT-2020 Task 2: Leveraging Transformers with Generalized Augmentation for Identifying Informativeness in COVID-19 Tweets}
\author{Nickil Maveli \\
  ILCC, School of Informatics \\
  University of Edinburgh \\
  \eml{n.maveli@sms.ed.ac.uk}}

\date{}

\begin{document}
\maketitle

\begin{abstract}
Twitter and, in general, social media has become an indispensable communication channel in times of emergency. The ubiquitousness of smartphone gadgets enables people to declare an emergency observed in real-time. As a result, more agencies are interested in programmatically monitoring Twitter (disaster relief organizations and news agencies). Therefore, recognizing the informativeness of a Tweet can help filter noise from the large volumes of Tweets. In this paper, we present our submission for \emph{WNUT-2020 Task 2: Identification of informative COVID-19 English Tweets}. Our most successful model is an ensemble of transformers, including RoBERTa, XLNet, and BERTweet trained in a Semi-Supervised Learning (SSL) setting. The proposed system achieves an $F_1$ score of  $\boldsymbol{0.9011}$ on the test set (ranking  $\boldsymbol{7}$\textsuperscript{th} on the leaderboard) and shows significant gains in performance compared to a baseline system using FastText embeddings.
\end{abstract}
\section{Introduction}
In late December 2019, there was an identification of an outbreak of a novel coronavirus causing COVID-19.\footnote{\url{https://www.ncbi.nlm.nih.gov/pmc/articles/PMC7159299/}} Due to the rapid spread of the virus,  the World Health Organization declared a state of emergency. Among several social media platforms, Twitter provides a powerful lens for identifying people’s behavior, decision-making, and sources of information before, during, and after wide-scope events, such as natural disasters~\citep{DBLP:conf/wsdm/BeckerNG10}. 
Due to the low signal-to-noise ratio, identifying relevant information in Tweets is a challenging task.

\begin{figure}[htbp!]
    \centering
    \includegraphics[width=2.5in, height=1.5in]{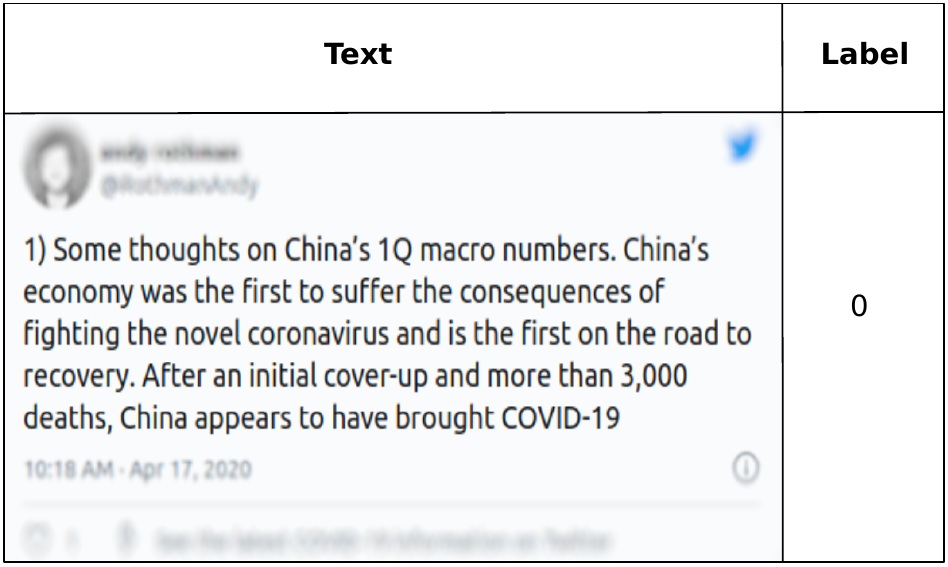}
    \caption{A hard to classify Tweet.}
    \label{fig:Tweet-sample}
\end{figure}

The basic goal of WNUT-2020 Task 2~\citep{covid19Tweet} is to automatically identify whether a COVID-19 English Tweet is Informative or not. Such Informative Tweets provide information about recovered, suspected, confirmed, and death cases as well as the location or travel history of the cases. About 4M COVID-19 English Tweets are daily being posted on Twitter, most of which are, however, not informative. In many scenarios, it is not always clear whether a person’s Tweet is announcing a disaster response. Consider an example of an \textsc{Uninformative} Tweet from the dataset as shown in Figure~\ref{fig:Tweet-sample}.

However, this observation is near inscrutable examining only the vocabulary
used; the Tweet contains a variety of top frequent informative words
({\textit{``coronavirus"}}, {\textit{``covid-19"}}, {\textit{``deaths"}}). This example hints
that in order to reach meaningful results, we may have to
examine contextual linguistic features, model the annotator's bias, introduce adversarial examples, and so on~\citep{DBLP:conf/emnlp/GevaGB19, DBLP:journals/corr/GoodfellowSS14}.
\begin{figure*}[t]
\centering
\includegraphics[height=9cm, width=15.4cm]{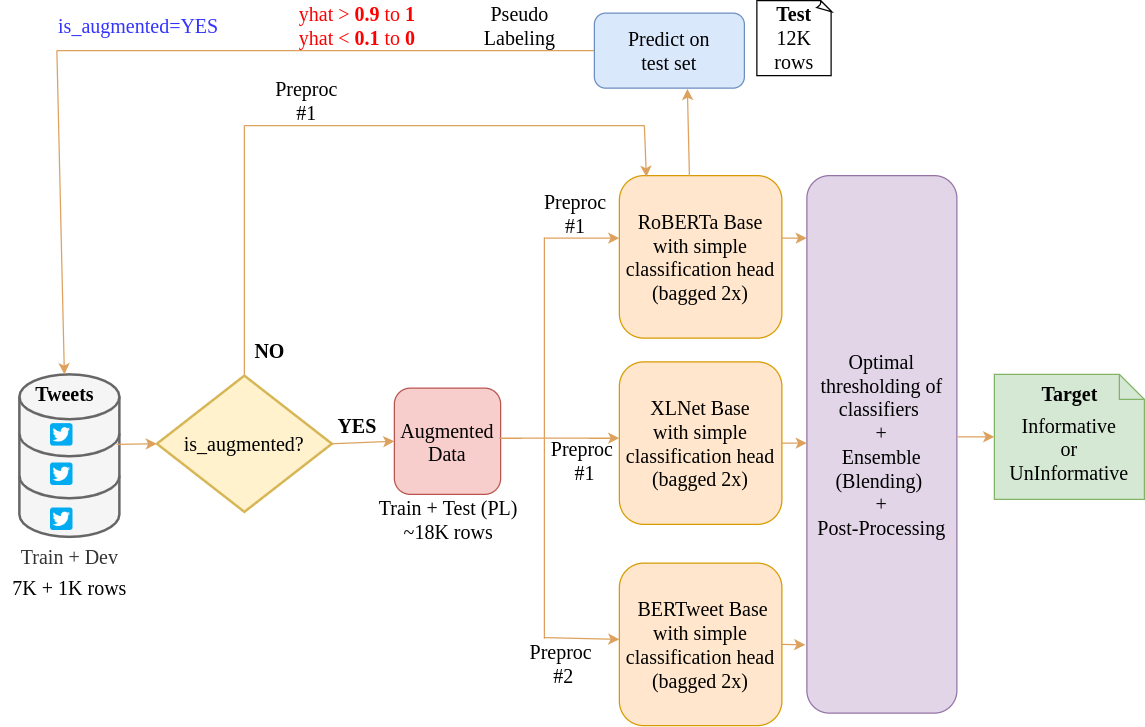}
\caption{Our proposed model architecture. A RoBERTa model does the classification on the 12K test-set, while being trained using 7K train-set. Later, 11K most confident predictions are appended to the train-set. The new concatenated data is fed to the ensemble models leading to better generalization and improved model performance.}\label{fig01:Architecture}
\end{figure*}

In this paper, we build an ensemble of Transformer \citep{DBLP:conf/nips/VaswaniSPUJGKP17} models to leverage its strength in capturing contextual information. The data used to train these models is an augmented version carefully designed to alleviate the confirmation bias and thereby improve generalization. The final inference result is the majority voting of the class from all the constituent models through optimal thresholding as a post-processing step. Our best model (ensemble) achieves $F_1$ scores of $\boldsymbol{0.9248}$ and $\boldsymbol{0.9011}$ on the development and test set respectively.
\section{Related Work}
Recently, research has started to investigate the use of deep learning in the area of disaster
response. For instance, \citet{Caragea2016IdentifyingIM} use CNN to detect informative messages in data from flood-related disasters and report significant improvements in performance over SVM and fully connected Neural Networks. \citet{DBLP:conf/icwsm/NguyenAJSIM17} use CNNs to capture the most salient
n-gram information on situational awareness crisis data and note the improvements over conventional algorithms. \citet{DBLP:conf/ifip12/LazregGG16} use LSTM network to learn a model from crisis Tweets to generate snippets of information for summarizing the Tweets. \citet{DBLP:conf/trec/WangL19} classify crisis-related Tweets using ELMo contextual word embeddings, whereas \citet{Ma2019TweetsCW} use a monolingual BERT-based model for Tweets classification problem in the disaster management field.

Text classification generally consists of two processes --- an encoder that converts textual inputs to numerical representations and a classifier that estimates hidden relations between these representations and class labels. The text representations are generated using N-gram statistics \citep{DBLP:conf/acl/WangM12}, word embeddings~\citep{DBLP:conf/eacl/GraveMJB17, DBLP:conf/acl/HenaoLCSWWZZ18}. More recently, powerful pre-trained models for text representations, e.g. BERT \citep{DBLP:journals/corr/abs-1810-04805}, have shown state-of-the-art performance on text classification tasks using only the simple classifier of a fully connected layer.
\section{System Description}
We formulate the task of identifying informativeness in Tweets as a binary text classification problem with \textsc{Informative} and 
\textsc{Uninformative} as the class names.
As shown in Figure \ref{fig01:Architecture}, the framework of our Informativeness classification model consists of three modules --- Transformer and BERTweet ensemble learning, generalized augmentation via pseudo-labeling, and optimal thresholding via post-processing to adjust the distribution of class labels in target. 

\subsection{Data Preprocessing}
The preprocessing pipeline consists of the following two strategies:
\begin{itemize}
  \item \textbf{Preproc \#1:}
  We lowercase the Texts and remove the Non-ascii letters, urls, @RT:[NAME], @[NAME]. Furthermore, we break apart common single tokens; Eg: RoBERTa makes a single token for \texttt{"..."}, so we convert all single \texttt{[...]} tokens into three \texttt{[.][.][.]} tokens. We split \texttt{"!!!"} in the same manner. All Transformer models use this preprocessing strategy.
  \item \textbf{Preproc \#2:}
  We normalize the Texts using \texttt{TweetTokenizer}.\footnote{\url{https://github.com/VinAIResearch/BERTweet/blob/master/TweetNormalizer.py}} Some of the normalization steps are --- Expand text contractions (\textit{``can't"} to \textit{``cannot"}, \textit{``M"} to \textit{``million"}, etc), text normalization (\textit{``p . m ."} to \textit{``p.m."}, etc). All BERTweet models use this preprocessing strategy.
\end{itemize}

\subsection{Model}
We train 6 models in total --- 2 each of RoBERTa-base, XLNet-base-cased, and BERTweet-base respectively on a 5-fold setup to find the optimal epoch. Its performance is evaluated on the validation set after every epoch. Later, it is trained on the complete dataset.

\subsubsection{RoBERTa} 
The meaning of words vary subtly across different contexts, and RoBERTa generates contextualized word representations to capture the context-sensitive semantics of words \citep{DBLP:journals/corr/abs-1907-11692}. The use of word representations from RoBERTa results in the state-of-the-art performance in a wide variety of language understanding tasks. Given a sentence $s$ consisting of $n$ words ${\{w_{1},\dots,w_{n}}\}$, RoBERTa model generates their contextualized representations ${\{\textbf{v}^{c}_{s,w_{1}}},\dots,{\textbf{v}^{c}_{s,w_{n}}\}}$.

\subsubsection{XLNet}
XLNet is an auto-regressive language model which is based on the transformer architecture with recurrence~\citep{DBLP:conf/nips/YangDYCSL19}. It outputs the joint probability of a sequence of tokens. The training objective calculates the probability of a word token conditioned on all permutations of word tokens in a sentence, as opposed to just those to the left or to the right of the target token.

\subsubsection{BERTweet} 
It is the first public large-scale language model pre-trained for English Tweets that is trained using a 80 GB corpus of 850M English Tweets \citep{BERTweet}. It uses the same architecture as BERT-base, which is trained with a Masked Language Modeling (MLM) objective \citep{DBLP:journals/corr/abs-1810-04805}. BERTweet-base model claims to do better than RoBERTa-base and outperforms previous SOTA models on three downstream Tweet NLP tasks of POS tagging, NER, and text classification.

\begin{table}[t]
\centering
\resizebox{0.8\columnwidth}{!}{%
\begin{tabular}{lll}
\toprule
\textbf{Parameter} & \textbf{Version 1} & \textbf{Version 2}\\
\midrule
\small{Max Sequence Length} & \small{128} & \small{192}\\
\small{Epochs} & \small{4} & \small{4}\\
\small{Batch Size} & \small{16} & \small{16}\\
\small{Learning Rate} & \small{2e-5} & \small{3e-5}\\
\small{Optimizer} & \small{Adam} & \small{AdamW (0.01)}\\
\small{FGM} & \small{no} & \small{yes}\\
\bottomrule
\end{tabular}%
}
\caption{\label{table:hyperparameters} Training Hyperparameters. To perform bagging, Version 1 and Version 2 are used.}
\end{table}

\subsubsection{Loss}
Training Loss, Binary Cross Entropy Loss is defined as follows:
\begin{align*}
\text{BCE} = \left\{\begin{matrix} & - \log(f(s_{1})) & & \text{if} & t_{1} = 1 \\ & - \log(1 - f(s_{1})) & & \text{if} & t_{1} = 0 \end{matrix}\right.
\end{align*}
where, $f()$ is the $sigmoid$ function and $s_1$ and $t_1$ are the score and the ground truth label for the class $C_1$, which is also the class $C_i$ in $C$. 

\begin{figure}[t]
\centering
\includegraphics[height=10.9cm, width=7.5cm]{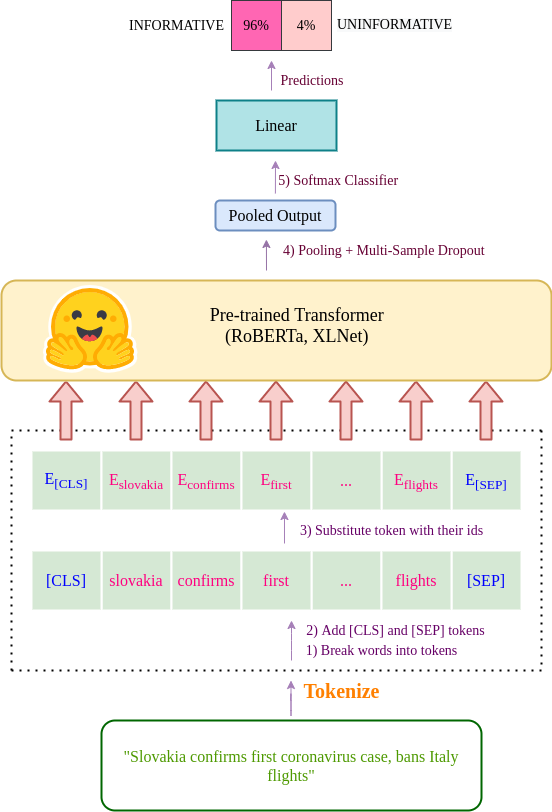}
\caption{Pre-trained Transformer model architecture for informativeness classification.}\label{fig02:Transformers}
\end{figure}

\begin{table}[!htbp]
\resizebox{\columnwidth}{!}{%
\begin{tabular}{lcccccc}
\toprule
\multicolumn{1}{c}{\textbf{Model}} & \multicolumn{3}{c}{\makecell{\textbf{Without} \\ \textbf{Augmentation}}} & \multicolumn{3}{c}{\makecell{\textbf{With} \\ \textbf{Augmentation}}}\\
\cmidrule(r){2-4}\cmidrule(l){5-7}
\multicolumn{1}{c}{} & {\textbf{Precision}} & \textbf{Recall} & $\boldsymbol{F_1}$ & \textbf{Precision} & \textbf{Recall} & $\boldsymbol{F_1}$\\
\midrule
{RoBERTa-base}\textsubscript{1} & 0.8652 & 0.9386 & 0.9004 & 0.9619 & 0.8833 & 0.9209\\ 
{RoBERTa-base}\textsubscript{2}  & 0.8760 & 0.9280 & 0.9012 & \textbf{0.9640} & 0.8818 & 0.9211\\ 
\midrule
{XLNet-base}\textsubscript{1}  & 0.8583 & 0.9364 & 0.8956 & 0.9619 & 0.8798 & 0.9190\\
{XLNet-base}\textsubscript{2}  & 0.8580 & 0.9343 & 0.8945 & 0.9619 & 0.8731 & 0.9153\\
\midrule
{BERTweet-base}\textsubscript{1}  & 0.8630 & 0.9343 & 0.8973 & 0.9534 & 0.8858 & 0.9184\\
{BERTweet-base}\textsubscript{2}  & 0.8483 & \textbf{0.9597} & 0.9006 & 0.9449 & 0.8974 & 0.9206\\
\midrule
\textbf{Ensemble} & 0.8790 & 0.9386 & 0.9078 & 0.9513 & 0.8998 & \textbf{0.9248}\\
\bottomrule
\end{tabular}%
}
\caption{\label{dev-results}Results on the Development set.}
\end{table}
\begin{table}[!htbp]
\centering
\resizebox{0.8\columnwidth}{!}{%
\begin{tabular}{cccc}
\toprule
\textbf{Model} & \textbf{P} & \textbf{R}  & $\boldsymbol{F_1}$\\
\midrule
\small{Baseline FastText} & \small{0.7730} & \small{0.7288} & \small{0.7503}\\
\midrule
\makecell{\small{RoBERTa-XLNet-}\\\small{BERTweet-Ensemble}} & \small{0.8768} & \small{0.9269} & \small{0.9011}\\
\bottomrule
\end{tabular}%
}
\caption{\label{table:test-results} Results on the Test set.}
\end{table}

\section{Experimentation}
We rely on the dataset provided by the organizers to perform our experiments. Overall, there are a total of 10K Tweets split in the ratio of 70/10/20 parts into train/development/test set respectively. However, for the final evaluation, 12K unlabeled noisy Tweets are provided, out of which 2K test Tweets are the actual ones the models are finally evaluated upon.
\subsection{Setup}
We install the huggingface \texttt{transformers} library \citep{DBLP:journals/corr/abs-1910-03771}. Pretrained RoBERTa-base and {XLNet-base-cased} models with a single linear layer which is simply a feed-forward network that acts as a classification head, are used. Figure~\ref{fig02:Transformers} shows a high-level overview of the architecture.

To speed up the training, sequence bucketing by removing unnecessary padding is employed \citep{DBLP:journals/corr/abs-1708-05604}.
To improve the robustness of neural networks, and improving resistance to adversarial attacks, Fast Gradient Method (FGM) is used at the end of the Transformer models~\citep{DBLP:conf/iclr/MiyatoDG17}. Multi-Sample Dropout \citep{DBLP:journals/corr/abs-1905-09788} is used when using dropout before the last layer with $p=0.5$, as it seem to converge the loss faster. We pass the output of each dropout layer to a shared weight fc layer. Lastly, we take the average of the outputs from fc layer as the final output. Table~\ref{table:hyperparameters} lists the chosen hyperparameters during model training.

For the {BERTweet-base} model, we normalize and tokenize Tweets with a {CNN}-{Dropout} layer for the inference part.\footnote{\url{https://www.kaggle.com/christofhenkel/setup-tokenizer}} Through a bunch of hyperparameters experimented from a finite sample space, we set the \text{batch\_size} = $16$, \text{epochs} = $5$, \text{max\_seq\_len} = $128$, \text{learning\_rate} = $3e-6$, along with Learning Rate Schedulers \citep{DBLP:conf/iclr/LoshchilovH17}.
\subsection{Augmentation}
We augment the data carefully with the help of pseudo labeling which is the process of adding confident predicted test data to the training data. Inorder to make the Cross Validation (CV) scheme less over-optimistic, we exclude the pseudo labels from the validation folds. In other words, once we retrieve the labels, we run the Kfold technique on only the original data points with real labels, and then add the labels to train exactly at training time. That way, the CV isn't biased by easy and artificially noiseless targets. It is augmented using the criteria:
\begin{align*}
\hat{y_{\text{new}}} = \left\{\begin{matrix} & 1 & & \text{if} & \hat{y_\text{roberta}} \geq 0.9 \\ & 0 & & \text{if} & \hat{y_\text{roberta}} < 0.1 \end{matrix}\right.
\end{align*}
where, $\hat{y_\text{roberta}}$ is the meta-prediction on the 12K test-set using {RoBERTa-base} and $\hat{y_{\text{new}}}$ is the new label associated with it. These are then concatenated back to the train set, making an augmented data of 18.915K Tweets to develop the final model. In other words, 11.915K out of 12K Tweets in the test-set are identified as confident predictions after pseduo labeling. The thresholds are decided based on several optimization ranges so as to maximize the $F_1$ score on the holdout development set.
\subsection{Post-Processing}
The idea here is to make the distribution of labels in development/test set to match corresponding distribution of labels in the train set so as to maintain the class ratio. Hence, the probabilities from all the 6 models are added and a majority voting cutoff value of $\boldsymbol{4}$ is found out by fine-tuning that maximize the $F_1$ score on the holdout development set.
\begin{align*}
p &= \sum_{i=1}^{6}\Vec{p_{i}}\\
p_{\text{out}} &= \left\{\begin{matrix} & 1 & & \text{if} & p \geq 4 \\ & 0 & & \text{if} & p < 4 \end{matrix}\right.
\end{align*}
where, $\Vec{p_{i}}$ is the probability vector calculated by the 6 models $i\in\{{1,\dots,6}\}$. $p$ is the ensemble output, whereas $p_{\text{out}}$ is the final prediction.
\section{Results and Error Analysis}
We conduct ablation analysis to compare the performance of our model variants. We evaluate the effect of contextual features by comparing our model with and without augmentation.
Table~\ref{dev-results} summarizes the performance on the development set.

Without the augmentation, we notice a situation of high Recall, low Precision. Our classifier thinks a lot of Tweets as belonging to the \textsc{Informative} class. This likely leads to a higher number of False Positive measurements, and a lower overall accuracy. For the {BERTweet-base\textsubscript{2}} model that gives the highest recall, 81 False Positive and 19 False Negative cases are identified. Whereas with the augmentation, a situation of low Recall, high Precision is observed. This makes sense as the model has access to more positive training samples and is able to make better decisions. Our classifier is very picky, and does not think many Tweets are \textsc{Informative}. For the {RoBERTa-base\textsubscript{2}} model that gives the highest Precision, 61 False Positive and 17 False Negative cases are identified. Ideally, in the real-world scenario, the high Recall case would be more favourable as we want the model to label everything that could potentially be an \textsc{Informative} Tweet, because a human personnel will most likely then interpret these results.

Understandably, the fine-tuned {RoBERTa} model did outperform every other experimented models. Bagging the models also lead to lower variance and robust predictions. Table~\ref{table:test-results} shows the final results on the Test set, wherein our model improves the organizer's baseline by 20\%. The effect of augmentation in the final ensemble is drastic as the $F_1$ score increases by about 1.87\%. 
Moreover, the idea of summing the probabilities of single models while ensembling worked better in comparison to choosing the most common label after finding different cutoff points that maximize the $F_1$ score of individual models. 

\begin{figure}
\centering
\includegraphics[height=5cm, width=5.5cm]{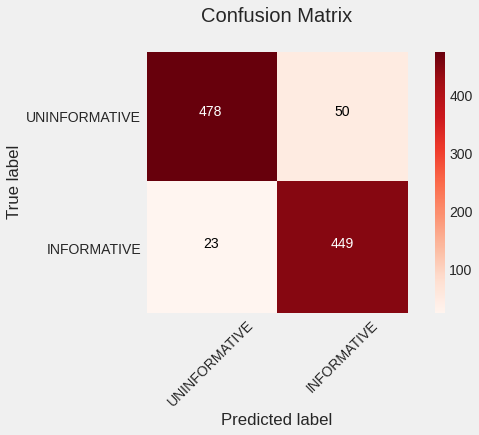}
\caption{Confusion Matrix.}\label{fig03:confusion-matrix}
\end{figure}

The confusion matrix of our best model is as shown in Figure~\ref{fig03:confusion-matrix}.
We look through the examples where our model made misclassification, and summarize the patterns of these error examples along with their attention visualization \citep{DBLP:conf/acl/Vig19}.
\begin{itemize}
    \item \textbf{Inaccurate interpretation of context:} In the sentence, \textit{``Writing 101: don’t put 2 numbers side by side. The punctuation is easy to miss. I first read this as being 51,385 people have died in Ontario from Covid."}, much of the attention weights are focused on the latter part. Our model may not capture this shift correctly given the long-distance dependency, which results in a False Positive prediction. See Figure \ref{fig04:attention_visualization_1} in Appendix \ref{sec:appendix} for attention visualization.
    \item \textbf{Misinformation due to ambiguity and subjectivity:} In the sentence, \textit{``I just remember this news recently China keeping two sets of coronavirus pandemic numbers? “Leaked” infection numbers over 154,000; deaths approach 25,000''}, it could be well evident that some events may not really happen as the source of the news lacked credibility. This could have prompted inter-annotator disagreement. See Figure \ref{fig05:attention_visualization_2} in Appendix \ref{sec:appendix} for attention visualization.
\end{itemize} 
\section{Conclusion}
We adopt an ensemble approach to reduce the variance of predictions and improve the model performance. The empirical results show the effectiveness and robustness of our model. Additionally, we perform a linguistic error analysis to gain insights into the model behavior. In the near future, we would like to combine user-related Tweet features (followers, friends, favorite counts) and Tweet-related meta-features (Retweets, creation date, sentiment) along with the contextual representation. Moreover, extending to multilingual Tweets \citep{DBLP:conf/acl/ChowdhuryCC20} is a potential future direction to pursue.

\section*{Acknowledgements}
We want to thank the anonymous reviewers for their time and comments which have helped make this paper and its contribution better.

\bibliographystyle{acl_natbib}
\bibliography{anthology,emnlp2020}

\appendix
\section{Appendix}
\label{sec:appendix}
\begin{figure}[H]
\centering
\includegraphics[height=15cm, width=5.5cm]{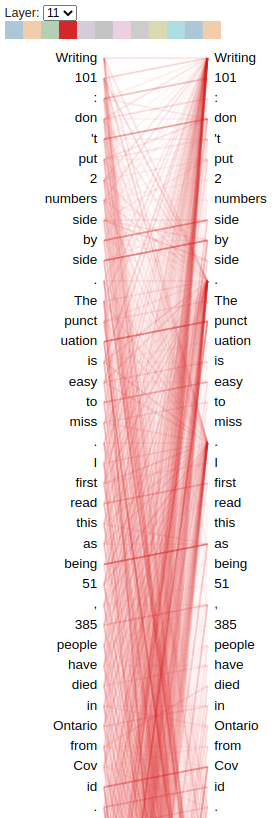}
\caption{Attention-head view for the last layer of RoBERTa-base showing attention to other words predictive of word. In this pattern, attention seems to be directed to other words that are predictive of the source word, excluding the source word itself. In the example below, most of the attention from “id" is directed to “Cov", whereas most of the attention from “Cov" is not focused on ``id".}\label{fig04:attention_visualization_1}
\end{figure}
\begin{figure}
\centering
\vspace*{-4cm}
\includegraphics[height=15cm, width=5.5cm]{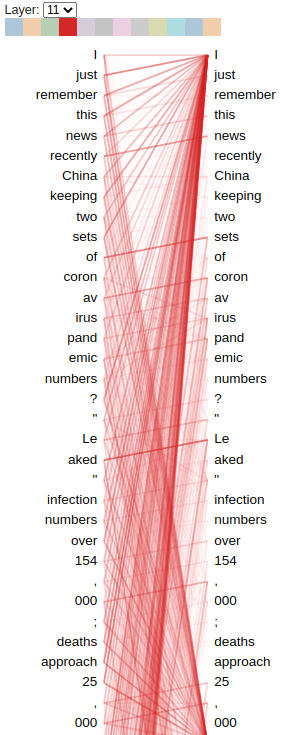}
\caption{Attention-head view for the last layer of RoBERTa-base showing attention to either the previous or the next token in the sentence. For instance, most of the attention for “China" is directed to the previous word “I". Considering a different example, most of the attention for “coron" is directed to the next word “irus" skipping “av" in between.}\label{fig05:attention_visualization_2}
\end{figure}
\end{document}